\documentclass[conference,a4paper]{IEEEtran}
\IEEEoverridecommandlockouts
\usepackage{cite}
\usepackage{amsmath,amssymb,amsfonts}
\usepackage{algorithmic}
\usepackage{graphicx}
\usepackage{textcomp}
\usepackage{xcolor}
\usepackage{booktabs}
\usepackage{microtype}

\def\BibTeX{{\rm B\kern-.05em{\sc i\kern-.025em b}\kern-.08em
    T\kern-.1667em\lower.7ex\hbox{E}\kern-.125emX}}
\begin{document}

\title{ Site-specific Deep Learning Path Loss Models based on the Method of Moments} 

\author{\IEEEauthorblockN{Conor Brennan}
\IEEEauthorblockA{\textit{School of Electronic Engineering} \\
\textit{Dublin City University}\\
Dublin, Ireland \\
conor.brennan@dcu.ie \\
https://orcid.org/0000-0002-0405-3869}
\and
\IEEEauthorblockN{Kevin McGuinness}
\IEEEauthorblockA{\textit{School of Electronic Engineering} \\
\textit{Dublin City University}\\
Dublin, Ireland \\
kevin.mcguinness@dcu.ie \\
https://orcid.org/0000-0003-1336-6477}
}

\maketitle

\begin{abstract}
This paper describes deep learning models based on convolutional neural networks applied to the problem of predicting EM wave propagation over rural terrain. A surface integral equation formulation, solved with the method of moments and accelerated using the Fast Far Field approximation, is used to generate synthetic training data which comprises path loss computed over randomly generated 1D terrain profiles. These are used to train two networks, one based on fractal profiles and one based on profiles generated using a Gaussian process.  The models show excellent agreement when applied to test profiles generated using the same statistical process used to create the training data and very good accuracy when applied to real life problems.  \end{abstract}


\section{Introduction}
The modelling of EM wave propagation over terrain is a central problem of wireless network design. The physical scale of the problem has meant that radio planners have historically relied on empirical curve-fitting  approaches \cite{Hata1980} or knife edge models\cite{Deygout1966}. More accurate formulations, based on for example, surface integral equations  (IE) exist \cite{Hviid1995}, but are slow if un-accelerated. A variety of acceleration techniques exist, most notably the Fast Far Field Approximation (FAFFA) \cite{Brennan1998b}. These computational efficiencies can be optimised using the Tabulated Interaction Method (TIM) \cite{Brennan1998_TIM,Brennan2015} which offers rapid, accurate simulation but is currently somewhat restricted in its formulation to relatively smooth surfaces. The purpose of this paper is to develop a model which, like the TIM, is both a) similar in accuracy to the FAFFA and b) an order of magnitude faster to implement. Crucially we seek to develop a model which is capable of being extended to more general problems in the future, an extension which is difficult for the TIM.   Machine learning (ML) potentially offers a framework to achieve this goal. It has been widely applied to propagation problems in recent years, including propagation in indoor \cite{Seretis2022,Bakirtzis2022} and urban \cite{Mao2022,Lee2019,Popescu2001,Gupta2022} scenarios. Rural deployments have also been considered such as in \cite{Moraitis2022, Ayadi2017,Egi2019} some based on relatively simple propagation models and others on the parabolic equation method \cite{Li2022}. 
 In this work we seek to develop an accurate {\em site-specific} model, that is one that can take in specific terrain height information as input and generate predictions for path loss along that particular profile.
 A key issue facing all ML techniques is access to  representative, accurate training data in sufficient quantity to produce reliable models.  This is particularly an issue in propagation modelling, where measured data is expensive in terms of hardware and man-hours. A popular alternative is to use synthetic data based on simulation, as was done in several of the works cited above. This is the  approach taken in this paper too, whereby propagation over tens of thousands of realistic profiles is efficiently analysed using the FAFFA. These data constitute  a training set which can then be used to develop an accurate, yet computationally  efficient, ML model. Synthetic data is justified in this instance on the reasonable basis that integral equation models have demonstrated very good agreement with measured data \cite{Hviid1995}.   In any event, several comparisons to measured data are presented  in section \ref{sec:results} so that the reader can gauge the performance.   The paper is organised as follows. Section  \ref{sec:EFIE} briefly describes the surface electric field integral equation and its efficient solution via the method of moments and  FAFFA algorithm. The process used to develop the data sets is presented in section \ref{sec:Dataset} while the development of the ML models is described in \ref{sec:ML}. Results are presented in section \ref{sec:results} and we close with conclusions and an outline of potential future work in section \ref{sec:conclusions}.

\section{EFIE Formulation}
\label{sec:EFIE}
In order to generate training data the path loss  over a set of artificially synthesised terrain profiles is computed.
The Electric Field Integral Equation (EFIE), solved with the method of moments,  is used to compute the path loss over each profile. 
\begin{figure}[htbp]
\centerline{\includegraphics[width=0.5\textwidth]{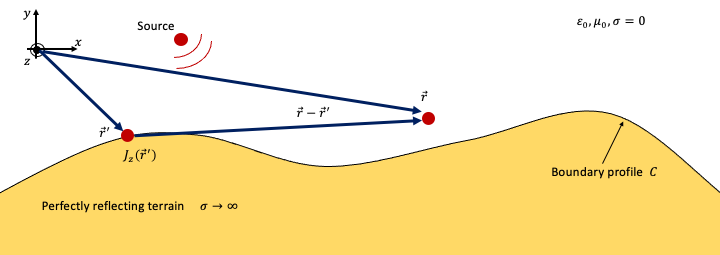}}
\caption{Geometry for the EFIE}
\label{fig:EFIE}
\end{figure}
Figure (\ref{fig:EFIE}) depicts a 2D problem where $TM^z$ incident fields emanate from a line source and impinge on a 1D surface.  A time variation of $e^{\jmath \omega t}$  is assumed and suppressed. 
For ease of implementation the terrain is assumed to be perfectly reflecting in this paper, a reasonable assumption at grazing incidence, but this is not a fundamental restriction and will be relaxed in future work. Under these assumptions the total field, $E_z^t$ at a general point $\bf r$ above the terrain surface can be written as 
\begin{equation}
E_z^{t} \left( \vec r \right) = E_z^s\left( \vec r \right) + E_z^i \left( \vec  r \right),
\label{eq:E_tot}
\end{equation}
where $E_z^i$ is the known incident field, i.e the field from the source that would exist in the absence of the scatterer while $E_z^s$ is the unknown scattered field caused by the presence of the terrain. The scattered field can be expressed in terms of an integral involving the surface electric current, $J_z$, as 
\begin{equation}
E_z^{s} \left( \vec r \right) = -\frac{k_0 \eta_0}{4} \int_C J_z \left( {\vec r}^\prime \right) H_0^{(2)} \left( k_0 \left| {\vec r } - {\vec r}^\prime \right| \right) dl^\prime,
\label{eq:E_sca}
\end{equation} where the integral takes place over the boundary of the scatterer (in this case the terrain surface) and $H^{(2)}_0$ is a zero-order Hankel function of the second kind. Applying the boundary condition of zero tangential fields for a point ${\vec r}$ on the scatterer surface yields the following equation for $J_z$
\begin{equation}
E_z^i \left( \vec r \right) = \frac{k_0 \eta_0}{4} \int_C J_z \left( {\vec r}^\prime \right) H_0^{\left( 2 \right)} \left( k_0 
\left| {\vec r } - {\vec r}^\prime \right| \right) ~dl^\prime.
\end{equation}
The method of moments is used to discretise the EFIE. 
The surface current is expanded using $N$ pulse basis functions, $f_n$, as 
\begin{equation}
    {J_z \left( \vec r \right)} \simeq \sum_{n=1}^N j_n f_n \left( {\vec r} \right),
\end{equation}
and point matching is applied at the basis domain centres to obtain a $N\times N$ dense linear system 
\begin{equation}
{\bf Zj } = {\bf v}.
\label{eq:matrix_eq}
\end{equation}
Equation (\ref{eq:matrix_eq}) can be solved in a variety of ways but it is computationally very expensive to do so when one considers that $N$ can be of the order of hundreds of thousands for a typical profile  at radio frequencies.   Assuming forward scattering (approximating $\bf Z$ as a lower-triangular matrix) 
allows (\ref{eq:matrix_eq}) to be solved using a straightforward, but slow, process of back substitution given by  
\begin{equation}
Z_{mm} j_m = V_m - \sum_{n<m} Z_{mn}j_n \mbox{ for } m = 1 \dots N.
\label{eq:back_sub}\end{equation}
Forward scattering is a reasonable assumption for the gently undulating terrain profiles considered in this paper, but nonetheless is an approximation that will be relaxed in future work.  
\subsection{FAFFA Acceleration}
\label{sec:FAFFA}
\begin{figure}[htbp]
\centerline{\includegraphics[width=0.5\textwidth]{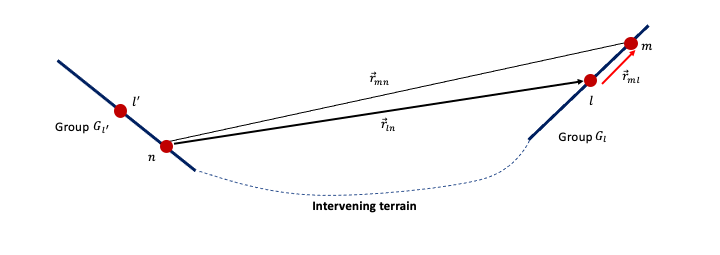}}
\caption{Geometry for the FAFFA}
\label{fig:FAFFA}
\end{figure}

Solution via  (\ref{eq:back_sub}) is effective but very slow and not suitable for the generation of training data which requires path-loss analysis for thousands of profiles.  To expedite the process the Fast Far Field Approximation (FAFFA) was implemented. The FAFFA  proceeds by assembling local collections of neighbouring pulse basis functions into $M$ groups. With such a decomposition (\ref{eq:back_sub}) can be equivalently written as 
\begin{eqnarray}
Z_{mm}j_m & = &  V_m - \sum_{l^\prime < l}\sum_{n \in G_{l^\prime}}Z_{mn}j_n -\sum_{n \in G_l, n < m }\hspace{-0.5cm} Z_{mn} j_n  
\label{eq:reordered_back_sub} \\
& & \mbox{ for } l = 1 \dots M, m \in  l.  \nonumber{}
\end{eqnarray}
The essence of the FAFFA is to replace the independent interactions between individual basis functions in separate groups with approximate interactions written in terms of a small number of computations that are extensively re-used. The approximation can be derived with reference to figure (\ref{fig:FAFFA}). 
Pulse basis functions $m$ and $n$ are situated in two groups, $G_l$ and $G_{l^\prime}$ with group centres $l$ and $l^\prime$ respectively.  A simple consideration of the geometry shows that
\begin{eqnarray}
\left| {\vec r}_{mn} \right| 
\simeq  
\left| {\vec r}_{ln} \right|  -  
 {\vec r}_{ml} \cdot {\hat r}_{ln}    \\
\simeq  
\left| {\vec r}_{ln} \right|  -  
 {\vec r}_{ml} \cdot {\hat r}_{ll^\prime},   
\end{eqnarray}  which for $\left| {\vec r}_{mn}\right|$  large allows the associated matrix element to be approximated as 
\begin{equation}
Z_{mn} \simeq Z_{ln} e^{\jmath k_0 {\vec r}_{ml} \cdot {\hat r}_{ll^\prime}   }. 
\end{equation}
Inserting this into  (\ref{eq:reordered_back_sub}) yields 
\begin{eqnarray}
Z_{mm} j_m & = &   V_m - \sum_{l^\prime < l}   e^{\jmath k_0 {\vec r}_{ml} \cdot {\hat r}_{ll^\prime}   }  \sum_{n \in G_{l^\prime}}Z_{ln}j_n \nonumber{} \\
& & -\sum_{n \in G_l, n < m }Z_{mn} j_n.  \label{eq:FAFFA_back_sub}
\end{eqnarray}
 for   $l = 1 \dots M, m \in  l$.
 The key computational advantage of the FAFFA is that for each pair of groups $G_l$ and $G_{l^\prime}$ once the summation $\sum_{n \in l^\prime}Z_{ln}j_n$, representing the fields scattered from group $G_{l^\prime}$ to the centre of group $G_l$, has been computed it can then be stored and repeatedly used to efficiently approximate the fields scattered from group $G_{l^\prime}$ to {\em every} point $m$ in group $G_l$. 
 In practice the groups are linear segments connecting sampled terrain heights. Consequently the final sum on the right hand side of  (\ref{eq:FAFFA_back_sub}), representing the interactions within a group, takes the form of a discrete convolution which can be efficiently computed using a Fast Fourier Transform, yielding an additional saving.  
Once the surface current has been computed using  (\ref{eq:FAFFA_back_sub}) total fields, and thus path-loss, at selected points above the terrain profile can be computed using  (\ref{eq:E_tot}) and (\ref{eq:E_sca}). Finally, in order to use this 2D simulation to make real-world predictions,  the total electric field at each point is multiplied by $\frac{1}{\sqrt{R}}$, where $R$ is the distance from the transmitter. This ensures that the power density decays as $\frac{1}{R^2}$ in free space, and serves as a heuristic conversion from a 2D field to a 3D field so as to best  compare to measured data.

\section{Machine Learning Data Set}
\label{sec:Dataset}
Synthetic data was created and used to develop two distinct ML models. In both cases  this involved randomly creating 8000 profiles and solving for the electric field at sampled locations above each profile.  In order to focus on an examination of  the ability of ML to model the physical scattering effects of the terrain we kept some parameters constant in each realisation. These parameters were frequency (assued to be 970$MHz$), transmitter height and location (10.4$m$ over the leftmost terrain point), and receiver height (2.4$m$ over the terrain at sampled points $50m$ apart).  Each profile comprised 256 sampled $(x,y)$ values where $x$ is range (in increments of 50$m$) and $y$ is height (chosen randomly in one of two ways, outlined below). The FAFFA was applied to efficiently solve for the fields at the sampled receiver locations.  Each of the 8000 elements of the data set thus comprised 256 $(x,y)$ points denoting the field point locations and a vector of 256 corresponding path loss values in dB. We developed two distinct ML models, one trained using synthetic profiles which were realisations of a Gaussian random process, while the second was trained using realisations created using  a fractal-generating algorithm. 

The first model, referred to as $ML_{GP}$, was based on a data set of size 8000 created using profiles which were realisations of a Gaussian random process. The root mean square height was set to 20$m$ while the correlation length was $800m$. 
The second model, referred to as $ML_F$ was based on a data set of size 8000 created using profiles which were random fractals created using the Diamond-Square algorithm \cite{Miller1986} with variance set to 30 and fractal parameter $H=1.2$. Some typical realisations (along with model validation results) for both profile types are shown in Fig.~(\ref{fig:test_gaussian}).

\section{ML model development}\label{sec:ML}
\begin{figure}[htbp]
\centerline{\includegraphics[width=0.3\textwidth]{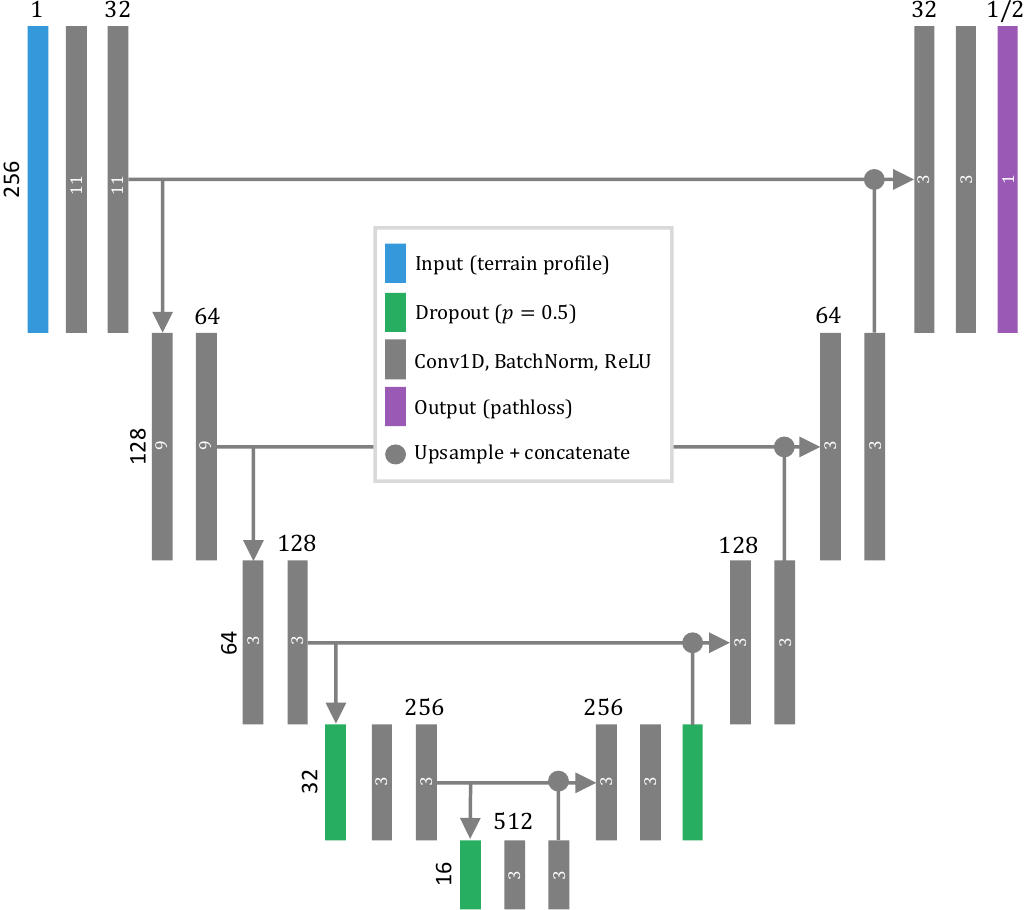}}
\caption{Network Architecture}
\label{fig:architecture}
\end{figure}

The goal of the machine learning model is to predict a $D$-dimensional path loss profile associated with a given $D$ dimensional terrain profile.  In this case $D$ refers to the number of $50m$ linear segments used to describe the profile. Here, we propose to use a deep neural network, $f_\theta : \mathbb{R}^D \to \mathbb{R}^D$, with parameters $\theta$ and train it using stochastic gradient methods on the synthetic training data. There are various model architectures that could be used for this, including convolutional neural networks (CNN), recurrent neural networks (RNN), and Transformer-based architectures. In this work we opted for a CNN based model, as unlike RNNs, these models can produce the entire path loss profile in a single pass and are often simpler to train than Transformer-based approaches.
The network architecture is shown in Fig. 
\ref{fig:architecture} and is based on U-Net \cite{ronneberger2015}, which is widely used in medical image and general semantic segmentation. The architecture is modified to handle 1D signals by replacing all 2D convolutional and batch normalization layers with their 1D counterparts. Three dropout layers ($p=\frac 12$) are also added to around the central bottleneck (after the 3rd and 4th downsampling layers and after the first upsampling layer) to provide regularization and reduce overfitting. Upsampling is done by linear interpolation without transposed convolutions. The output layer of the network produces a single scalar value for each input and no activation function is used on this layer.
We reduce the number of parameters in the network by approximately 50\% by halving the number of channels on the internal layers when compared to the original U-Net model. We also widen the kernel size on the initial two convolution layers from 3 to 11 to provide more spatial context. 
\subsection{Adam Optimiser, calibration and data augmentation}
The network was trained using a mean squared error loss function for 25 epochs over the training data using the Adam optimiser \cite{Kingma2015} with an initial learning rate of $10^{-2}$ and a weight decay $\lambda=10^{-5}$. The batch size was set to 128. The model is evaluated on the validation set after each epoch and the model with the lowest validation loss is retained for testing. The learning rate is stepped down by a factor of 10 when the validation loss has not reduced for 10 epochs. 
The initialization and batch normalization layers in neural networks are typically calibrated so that the output layer produces values that are approximately distributed according to a standard normal distribution before training. If left unchanged, this will cause the loss value to be very large in the early epochs, as the average value of the target is approximately -134 dB. To compensate, we set the bias value of the output layer to -134dB, which makes the network have an error close to that of a regressor that always produces the mean value (around 630 dB$^2$). This accelerates training since the network does not need to spend a long time during training moving the network predictions towards the range and scale of the targets.
Since the path loss is independent of the absolute height of the terrain profile and source/receiver points,  (in that it only depends  on the relative distance between the totality of  points), the network should produce a result that is invariant to changes in the absolute height of the input profile. To encourage this, we perform random data augmentation at training time to modify the absolute height of the input profile (keeping the relative vertical displacement of the source and receivers fixed at 10.4$m$ and $2.4m$ respectively). Specifically, we add a random scalar $\epsilon \sim \mathcal N(0, 30)$ to each input before passing it through the network. Other possible approaches to making the network produce outputs that are invariant to profile height shifts are to either normalize the profile to always start at height 0, or to use the profile derivatives as inputs instead of the profile heights. Empirically, however, we found the data augmentation approach more effective, likely because this approach also provides a degree of additional regularization.

\subsection{Uncertainty prediction}
\label{sec:uncertainty}
In addition to a point prediction of the path loss, in some applications it is useful to also provide an estimation of the uncertainty of the predictions. To facilitate this, we also trained a variation of the networks that also estimates the variance by modifying the network to produce two outputs:
\begin{equation}
    \mu_\theta(\mathbf x) = f_\theta(\mathbf x)_1, \quad \log\sigma^2_\theta(\mathbf x) = f_\theta(\mathbf x)_2,
\end{equation}
where now $f_\theta : \mathbb{R}^D = \mathbb{R}^{D\times 2}$ and the model predicts the log of the variance to ensure that the predicted variance is always positive. Assuming that the distribution of the target conditioned on the terrain profile $p(\mathbf y\mid \mathbf x) \sim \mathcal N(\mu_\theta(\mathbf x), \sigma_\theta(\mathbf x)I)$ gives a negative log likelihood loss of:
\begin{equation}
  -\log p(\mathbf y\mid \mathbf x) = \frac 12 \sum_{k=1}^D  \left(\frac{(y_k - \mu_\theta(\mathbf x)_k)^2}{\sigma^2_\theta(\mathbf x)_k} + \log \sigma^2_\theta(\mathbf x)_k\right).
\end{equation}
This variant of the model is again trained to minimize the expected value of the above loss using stochastic estimates on batches of size 128. It takes 3 times longer to train than the version that only produces point estimates (75 epochs). 

\subsection{Computation times and Validation}
Both model variants are relatively fast to train, requiring less than 5 minutes (NVIDIA GeForce RTX 3090). Inference takes approximately 2ms per profile on both GPU and CPU hardware (AMD Ryzen 9 5950x). Inference can also be done in batches, leading to approximately 10-20$\times$ better throughput in our experiments (batch size 128).
In each case the 8000 profiles were randomly split into 7500 for model development and 500 for validation. 
\label{sec:testing}
\begin{figure}[htbp]
\centerline{\includegraphics[width=0.5\textwidth]{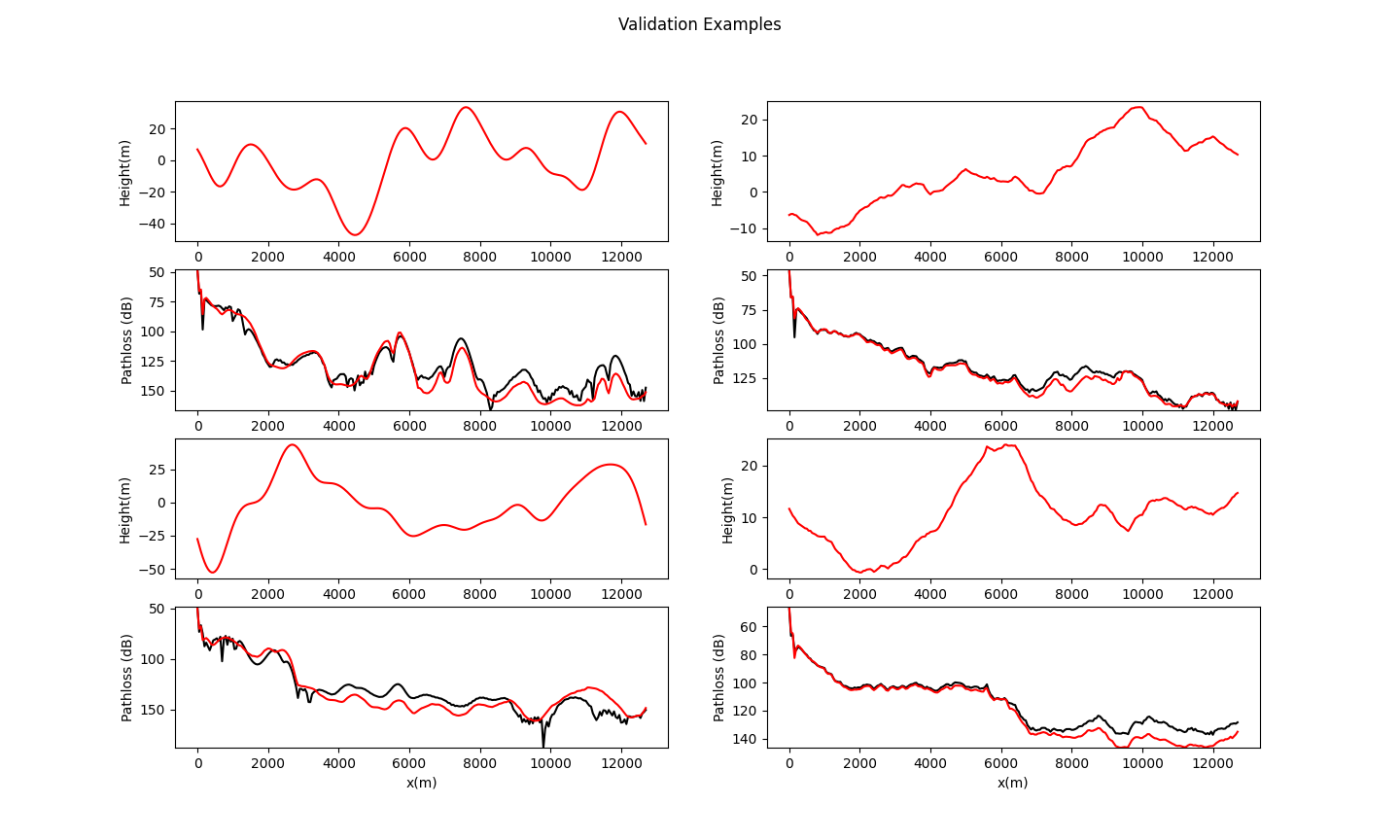}}
\caption{Validation examples for $ML_{GP}$(left) and $ML_F$(right). In each case profile is on top (red) while FAFFA (black) and $ML_F,ML_{GP}$ (red) pathloss predictions are below.}
\label{fig:test_gaussian}
\end{figure}

Fig. (\ref{fig:test_gaussian}) shows some randomly selected validation tests. Two examples with Gaussian profiles are shown on the left while the right shows two examples of fractal profiles.   The results show the ability of the ML model to accurately solve problems involving profiles of the same type. 
\section{Results}
\label{sec:results}
\begin{figure}[htbp]
\centerline{\includegraphics[width=0.5\textwidth]{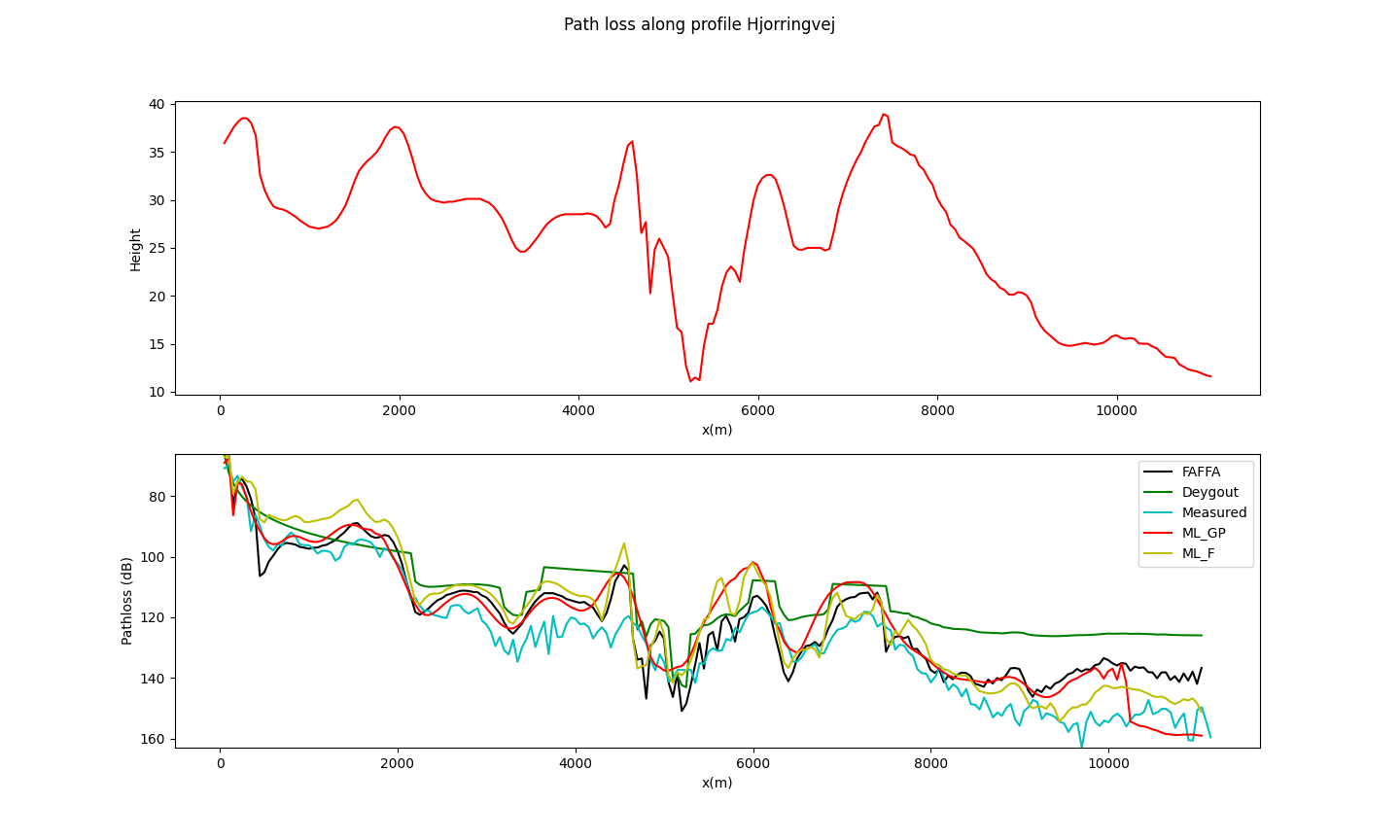}}
\centerline{\includegraphics[width=0.5\textwidth]{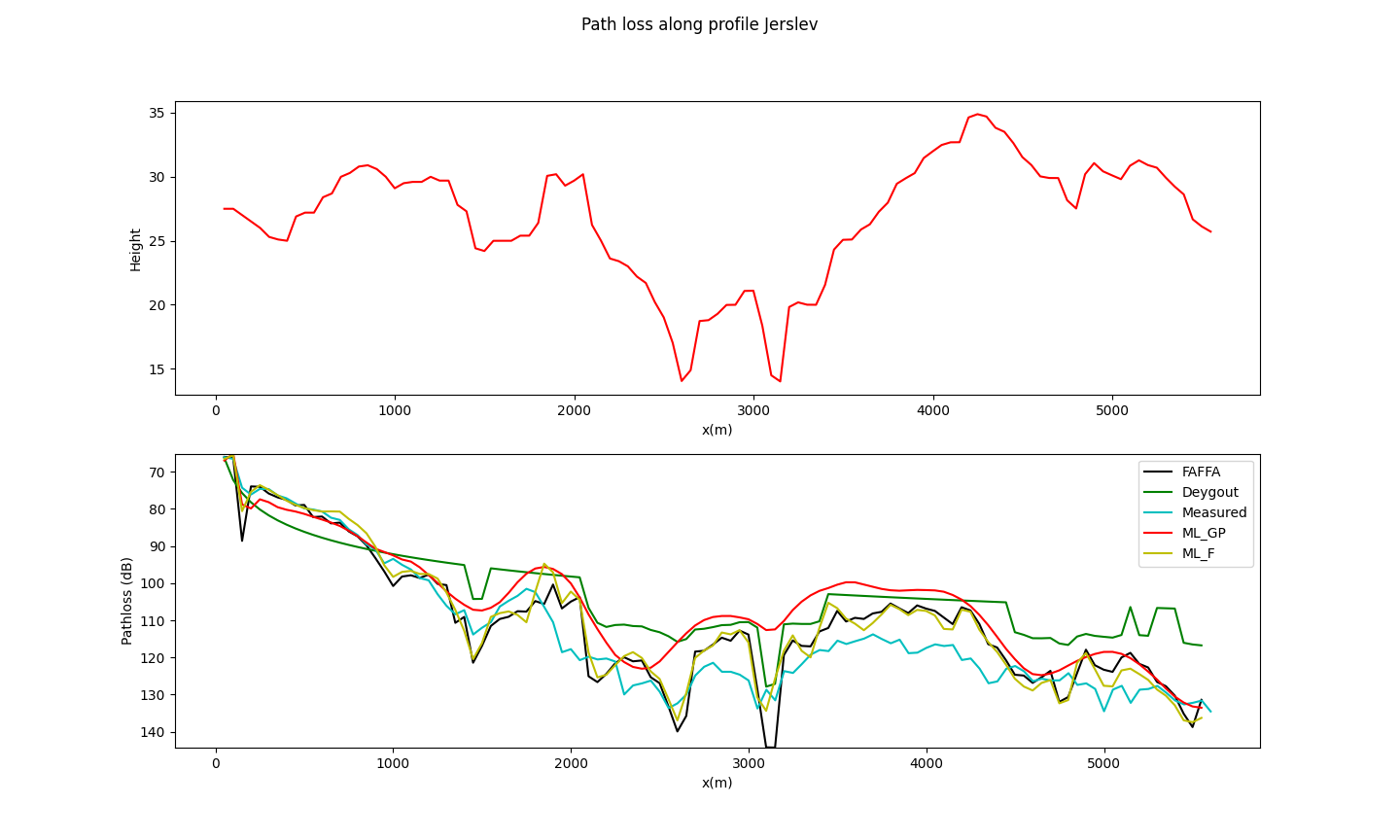}}
\caption{Profile and Path loss for Hjorringvej (top 2) and Jerslev (bottom 2).}
\label{fig:real_hjo}
\end{figure}

The validation results of section \ref{sec:testing} confirm that the ML model can rapidly reproduce the predictions of a full-wave EM solver with good accuracy when applied to profiles of a similar type (i.e. Gaussian or fractal).   To be of use in radio planning the models must generalise and be applicable to more general real-world profiles.  To do this we ran the trained models for several profiles for which measured data was available.  The results are shown in Figs (\ref{fig:real_hjo}). 
The terrain profile is shown on top while the bottom plot shows the measured path gain and predictions using the FAFFA, the two ML models and a knife edge model based on Deygout's method. Good agreement for the two ML models is noted with $M_F$, the model based on fractal data, outperforming $M_{GP}$ the model based on Gaussian profiles. This is perhaps un-surprising given the fractal-like appearance of real terrain profiles.  Table \ref{table:stats} provides statistics  about the performance of the models. $\overline{\epsilon}_F$ and $\sigma_F$ are average error and standard deviation of ML predictions relative to the FAFFA predictions, while  $\overline{\epsilon}_M$ and $\sigma_M$ are statistics (for all models) based on a comparison to the measured data.
In the context of the ML models the most  important data are arguably $\overline{\epsilon}_F$ and $\sigma_F$.   This is because  the ML models are trained on the FAFFA model and not the measurements. The FAFFA prediction thus serves as the upper-bound on their performance and their ability to match the measured data is thus limited to the performance of the FAFFA model.  
Any enhanced agreement with measurements over that of FAFFA, while superficially welcome,  would not necessarily be reproduced for other profiles and could not be considered indicative of an enhanced predictive capability.
Nonetheless it should be noted that, when compared to the measurements, both ML models have accuracy which is similar to the FAFFA, while having a greatly reduced  computational burden which is similar to the knife edge model.  
\begin{table}[!htbp]
\centering
\caption{Model Error Statistics}
\begin{tabular}{c|cccc|cccc}
\toprule
Model &  \multicolumn{4}{c}{Hjorringvej} & \multicolumn{4}{c}{Jerslev} \\
\midrule
{}   & $\overline{\epsilon}_F$   & $\sigma_F$    & $\overline{\epsilon}_M$   & $\sigma_M$ & $\overline{\epsilon}_F$   & $\sigma_F$    & $\overline{\epsilon}_M$   & $\sigma_M$   \\
ML$_F$ & 1.6 & 6.49 & 7.41& 6.15& 0.51& 3.19 & 3 & 5.68   \\
ML$_{GP}$ & 0.34 & 7.66& 6.15 & 6.68& 5.14 & 6.4 &7.63  &7.11   \\
FAFFA & - & - & 5.81 & 7.42 & - & - & 2.49 & 6.39   \\
Deygout & - & - & 14.11 & 9.41 & - & - & 9.79 & 8.03   \\
\bottomrule
\end{tabular}
\label{table:stats}
\end{table}

\begin{figure}[htbp]
\centerline{\includegraphics[width=0.5\textwidth]{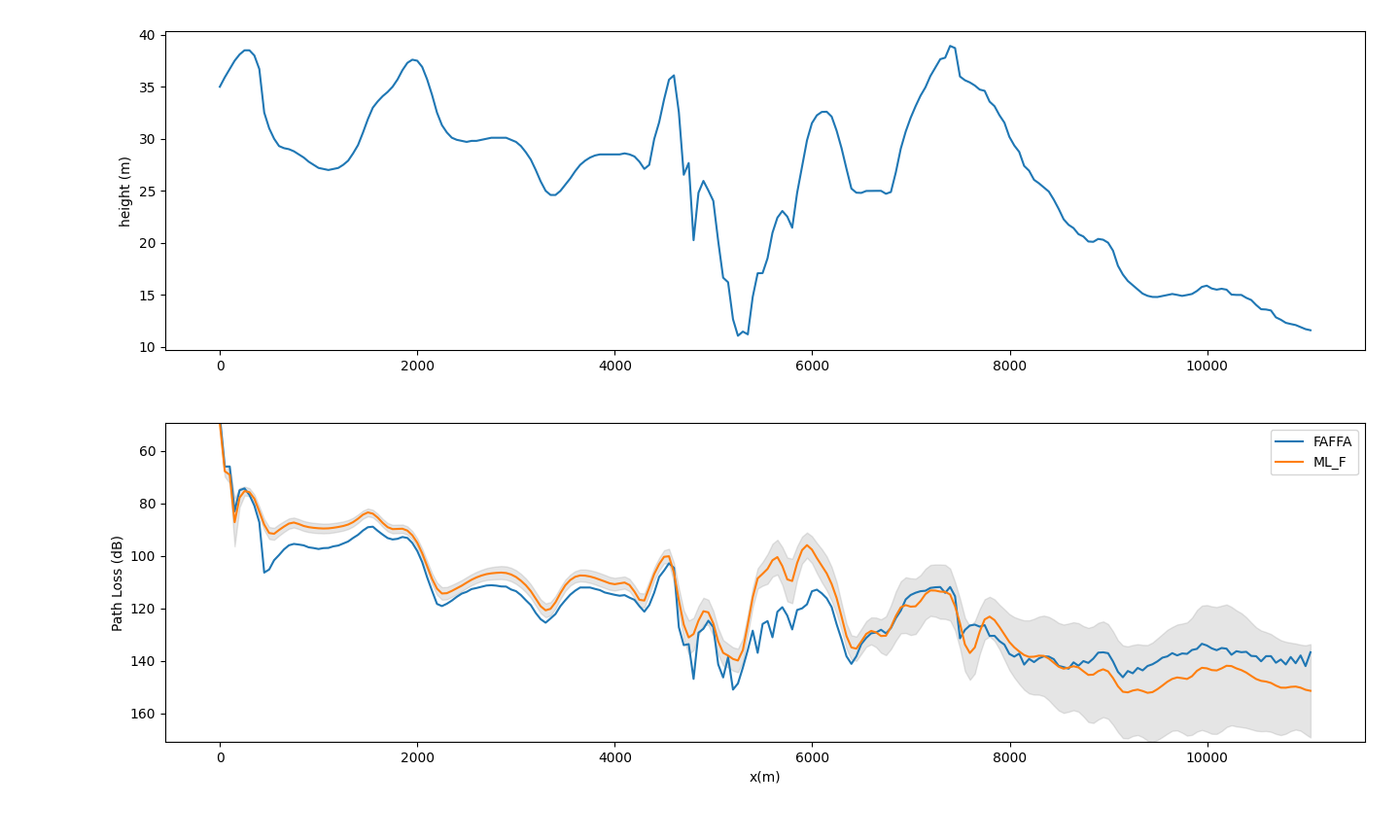}}
\caption{$ML_F$ prediction (red) with uncertainty regions (grey) for Hjorringvej profile. FAFFA prediction in blue. }
\label{fig:hjo_interval}
\end{figure}

Finally Fig. (\ref{fig:hjo_interval}) shows an example of the $ML_F$ prediction for a real profile, where the uncertainty interval is shown in grey (extending to two standard deviations around the $ML_F$ prediction in red). The computation of the model variance in the model is discussed in section (\ref{sec:uncertainty}). 

\section{Conclusions and future work} 
\label{sec:conclusions} 
This paper has presented two ML models for predicting path loss over rural terrain. The ML models are based on deep learning convolutional networks trained with synthetic data. The training data was obtained by generating path loss over thousands of randomly generated profiles, which were generated using either a Gaussian process or a fractal-generating algorithm. The training path loss data was obtained using the FAFFA, an accelerated method of moments solver. In both cases the agreement between the ML models and the validation data was excellent. When applied to real life profiles the model trained on fractal profiles was more accurate with both ML models outperforming a knife edge model.  Future work will concentrate on identifying a more representative data set and developing a more general model (which will be valid for multiple frequencies, arbitrary transmitter and receiver heights etc).  

\bibliographystyle{ieeetr}
\bibliography{EuCAP23_Bren_McG}

\begin{thebibliography}{10}

\bibitem{Hata1980}
M.~Hata, ``Empirical formula for propagation loss in land mobile radio
  services,'' {\em IEEE Transactions on Vehicular Technology}, vol.~29,
  pp.~317--325, 1980.

\bibitem{Deygout1966}
J.~Deygout, ``Multiple knife-edge diffraction of microwaves,'' {\em IEEE
  Transactions on Antennas and Propagation}, vol.~14, pp.~480--489, 1966.

\bibitem{Hviid1995}
J.~T. Hviid, J.~B. Andersen, J.~Toftgård, and J.~Bøjer, ``Terrain-based
  propagation model for rural area—an integral equation approach,'' {\em IEEE
  Transactions on Antennas and Propagation}, vol.~43, pp.~41--46, 1995.

\bibitem{Brennan1998b}
C.~Brennan and P.~J. Cullen, ``Application of the fast far-field approximation
  to the computation of uhf pathloss over irregular terrain,'' {\em IEEE
  Transactions on Antennas and Propagation}, vol.~46, pp.~881--890, 1998.

\bibitem{Brennan1998_TIM}
C.~Brennan and P.~Cullen, ``Tabulated interaction method for uhf terrain
  propagation problems,'' {\em IEEE Transactions on Antennas and Propagation},
  vol.~46, pp.~738--739, 1998.

\bibitem{Brennan2015}
C.~Brennan, D.~Trinh, and R.~Mittra, ``Two-level tabulated interaction method
  for electromagnetic scattering from lossy irregular terrain profiles
  incorporating back scattering,'' {\em IEEE Transactions on Antennas and
  Propagation}, vol.~63, pp.~4024--4036, 2015.

\bibitem{Seretis2022}
A.~Seretis and C.~D. Sarris, ``A hybrid machine learning-based model for indoor
  propagation; a hybrid machine learning-based model for indoor propagation,''
  in {\em 2022 16th European Conference on Antennas and Propagation (EuCAP)},
  2022.

\bibitem{Bakirtzis2022}
S.~Bakirtzis, K.~Qiu, J.~Zhang, and I.~Wassell, ``Deepray: Deep learning meets
  ray-tracing; deepray: Deep learning meets ray-tracing,'' in {\em 2022 16th
  European Conference on Antennas and Propagation (EuCAP)}, 2022.

\bibitem{Mao2022}
K.~Mao, Q.~Zhu, M.~Song, H.~Li, B.~Ning, G.~F. Pedersen, and W.~Fan, ``Machine
  learning-based 3d channel modeling for u2v mmwave communications,'' {\em IEEE
  Internet of Things Journal}, vol.~9, pp.~17592--17607, 2022.

\bibitem{Lee2019}
J.~G. Lee, M.~Y. Kang, and S.~C. Kim, ``Path loss exponent prediction for
  outdoor millimeter wave channels through deep learning,'' in {\em IEEE
  Wireless Communications and Networking Conference, WCNC}, 2019.

\bibitem{Popescu2001}
I.~Popescu, I.~Nafornita, P.~Constantinou, A.~Kanatas, and N.~Moraitis,
  ``Neural networks applications for the prediction of propagation path loss in
  urban environments,'' in {\em IEEE Vehicular Technology Conference}, 2001.

\bibitem{Gupta2022}
A.~Gupta, J.~Du, D.~Chizhik, R.~A. Valenzuela, and M.~Sellathurai, ``Machine
  learning-based urban canyon path loss prediction using 28 ghz manhattan
  measurements,'' {\em IEEE Transactions on Antennas and Propagation}, vol.~70,
  pp.~4096--4111, 2022.

\bibitem{Moraitis2022}
N.~Moraitis, L.~Tsipi, D.~Vouyioukas, A.~Gkioni, and S.~Louvros, ``On the
  assessment of ensemble models for propagation loss forecasts in rural
  environments,'' {\em IEEE Wireless Communications Letters}, vol.~11,
  pp.~1097--1101, 5 2022.

\bibitem{Ayadi2017}
M.~Ayadi, A.~B. Zineb, and S.~Tabbane, ``A uhf path loss model using learning
  machine for heterogeneous networks,'' {\em IEEE Transactions on Antennas and
  Propagation}, vol.~65, pp.~3675--3683, 7 2017.

\bibitem{Egi2019}
Y.~Egi and C.~E. Otero, ``Machine-learning and 3d point-cloud based signal
  power path loss model for the deployment of wireless communication systems,''
  {\em IEEE Access}, vol.~7, pp.~42507--42517, 2019.

\bibitem{Li2022}
A.~Li, C.~Yin, and Q.~Zhang, ``Predicting spatial field values under undulating
  terrain with 2w-pe based on machine learning,'' {\em IEEE Antennas and
  Wireless Propagation Letters}, vol.~21, 2022.

\bibitem{Miller1986}
G.~S. Miller, ``Definition and rendering of terrain maps.,'' {\em Computer
  Graphics (ACM)}, vol.~20, 1986.

\bibitem{ronneberger2015}
O.~Ronneberger, P.~Fischer, and T.~Brox, ``U-net: Convolutional networks for
  biomedical image segmentation,'' vol.~9351, 2015.

\bibitem{Kingma2015}
D.~P. Kingma and J.~L. Ba, ``Adam: A method for stochastic optimization,'' in
  {\em 3rd International Conference on Learning Representations, ICLR 2015 -
  Conference Track Proceedings}, 2015.

\end{thebibliography}
\end{document}